\documentclass[conference]{IEEEtran}
\ifCLASSINFOpdf
\else
\fi
\usepackage{amsmath}  
\usepackage{amssymb}  
\usepackage{amsfonts}
\usepackage{graphicx}
\usepackage{multirow}
\usepackage{amsmath}
\usepackage{float}
\usepackage{url}
\usepackage{booktabs}
\begin{document}
\title{Personalized Class Incremental Context-Aware Food Classification for Food Intake Monitoring Systems\vspace{-10pt}}
\author{\IEEEauthorblockN{ Hassan Kazemi Tehrani, Jun Cai, Abbas Yekanlou, and Sylvia Santosa}
\IEEEauthorblockA{Network Intelligence and Innovation Lab ($NI^2L$), Department of Electrical and Computer Engineering,\\
Concordia University, Montreal QC H3G 1M8, Canada \\
Department of Health, Kinesiology, and Applied Physiology,\\ SP165.21
Concordia University
7141 Sherbrooke Street West
Montreal, QC H4B 1R6, Canada\\
Emails: hassan.kazemitehrani@concordia.ca; jun.cai@concordia.ca; abbas.yekanlou@concordia.ca; s.santosa@concordia.ca}
\vspace{-30pt}
}

\maketitle
\begin{abstract}
Accurate food intake monitoring is crucial for maintaining a healthy diet and preventing nutrition-related diseases. In practical scenarios, with the diverse range of foods consumed across various cultures and regions, classic food classification models have significant limitations due to their reliance on fixed-sized food datasets. Moreover, studies show that people consume only a small range of foods across the existing ones, with each individual consuming a unique set of foods. These limitations necessitate the model to adapt itself as new classes of foods appear. Additionally, the model needs to pay more attention to certain food classes. Existing class-incremental models have low accuracy for the new classes and lack personalization. This paper introduces a personalized, class-incremental food classification model designed to overcome these challenges and improve the performance of food intake monitoring systems. Our approach dynamically adapts itself to the new array of real-world food classes, maintaining applicability and accuracy, both for new and existing classes by using personalization.
Our model's primary feature is personalization, which improves classification accuracy by prioritizing a subset of foods based on an individual's eating habits, including meal frequency, meal times, and locations of particular food classes.
To achieve a class-incremental setting, a modified version of the dynamic support network (DSN) is utilized to expand on the appearance of new food classes. Additionally, we propose a comprehensive framework that integrates this model into a food intake monitoring system. This system analyzes meal images provided by users, makes use of a smart scale to estimate food weight, utilizes a nutrient content database to calculate the amount of each macro-nutrient, and creates a dietary user profile through a mobile application. Finally, experimental evaluations on two new benchmark datasets FOOD101-Personal and VIPER-FoodNet-Personal (VFN-Personal), personalized versions of well-known datasets for food classification, are conducted to demonstrate the effectiveness of our model in improving the classification accuracy of both new and existing classes, addressing the limitations of both conventional and class-incremental food classification models.
\end{abstract}
\section{Introduction}
In recent years, having a healthy diet to improve the quality of life has become more ubiquitous. In order to maintain a healthy life and prevent nutrition-related diseases including diabetes and heart disease, accurately estimating daily nutritional intake is crucial, which can help control weight, manage diseases and boost physical and mental health. New technologies and tools like wearable devices have enabled food intake monitoring systems to track dietary habits, analyze nutritional information, and provide dietary advice \cite{singla2016food}. 

Due to the advances in smartphones and mobile technologies, individuals tend to take pictures of their meals and snacks more often, which increases the number of food images and the rise of demand for food image classification systems \cite{csengur2019food}.
Food image classification is a subcategory of image classification that involves using computer vision to categorize or label food images based on their contents \cite{raghavan2024online}. Existing works have shown remarkable outcomes on fixed-
sized datasets \cite{he2023single}, however, in multicultural urban settings where food
diversity is at its peak, these models generally struggle to adapt to new food categories not present in their fix training datasets. Consequently, there is a pressing need for models that can dynamically adapt and incorporate new food items continuously, a process referred to as class-incremental learning \cite{hsu2018re}.

Class-incremental learning models, which learn new classes without forgetting the previous ones, provide a viable solution for addressing the drawbacks of conventional methods. However, while state-of-the-art methods achieve good overall accuracy, their performance on incremental classes is noticeably low \cite{zhao2024bias}. As more frequent and user-specific food classes are the ones that appear incrementally, this limitation can hinder the goal of attaining precise dietary monitoring.

To address these challenges, this paper proposes a novel personalized class-incremental food classification model that enhances the adaptability and accuracy of food intake monitoring systems. Despite recent advancements, integrating contextual factors such as meal times, locations, and specific eating habits effectively remains a significant gap, hindering the achievement of universal accuracy across diverse dietary preferences. We proposed a personalized class incremental food classification model, a modified version of dynamic support network (DSN) \cite{yang2022dynamic}, that learns the new food classes not only efficiently but also with high accuracy. Through a focus on personalization, we aim to highlight the subset of foods most relevant to an individual’s diet, taking into account factors such as meal times, locations, and specific eating habits, thereby improving the classification accuracy across diverse eating patterns. Finally, we evaluated our model on two new benchmark datasets for personalized food classification, FOOD101-Personal and VIPER-FoodNet-Personal (VFN-Personal) \cite{pan2023personalized}, to show the effectiveness of our model. Experimental results show that the proposed approach not only improves the accuracy of classifying both new and existing food items but also enhances the applicability of the system in real-world scenarios.

In summary, the main contributions of this study include:
\begin{itemize}
\item We propose a novel personalized class-incremental food classification model able to achieve high accuracy for both existing and new classes.
\item We design a framework for food intake monitoring systems that leverages our personalized model as its core component.
\item We evaluated and compared our work with existing baseline models using new benchmark datasets, FOOD101-Personal and VFN-Personal.
\end{itemize}

\section{Related Work}
\label{sec:related}
\subsection{Food image classification}
The increase in available food datasets has further advanced the field of food image classification. Identifying food types directly from images is highly valuable for a range of food-related applications \cite{min2023large}. According to \cite{subhi2019vision}, several traditional machine learning methods are capable of classifying food images such as support vector machines (SVM) \cite{cortes1995support}, k-nearest neighbors (KNN) \cite{cover1967nearest}, 
and random forests \cite{breiman2001random}. These traditional methods use handcrafted features including color, texture and scale-invariant feature transform \cite{lowe2004distinctive}. Several papers have explored these techniques in the context of food image classification, each contributing unique 
perspectives and advancements \cite{joutou2009food}.
However, food naturally comes in a wide variety of looks, which contributes to high intra-class diversity: various foods within the same category might have quite diverse appearances. On the other hand, there is high inter-class similarity. Therefore, dietary groups may seem to be comparable to one another. These qualities put traditional techniques to the test since they frequently fail to identify the intricate details required for precise food identification \cite{yadav2021automated}.

Convolutional neural networks (CNNs) excel at image classification. CNNs automatically learn features through 2D convolutional layers, eliminating the need for manual feature extraction. This allows CNNs to perform well in detailed feature detection and makes them particularly accurate for computer vision tasks \cite{konstantakopoulos2023review}. There are some famous CNN architectures like AlexNet \cite{krizhevsky2012imagenet}, VGGNet \cite{simonyan2014very}, GoogLeNet (Inception v1) \cite{szegedy2015going}, ResNet \cite{he2016deep}, DenseNet \cite{huang2017densely}, and EfficientNet \cite{tan2019efficientnet} ordered by publication date respectively, which can be used for image classification tasks including food image classification. Multiple pieces of research have been conducted to utilize CNNs to classify food images with various network architectures and types of food datasets 
\cite{singla2016food}.
\begin{figure*}[ht!] 
\centering
\includegraphics[width=5.8in]{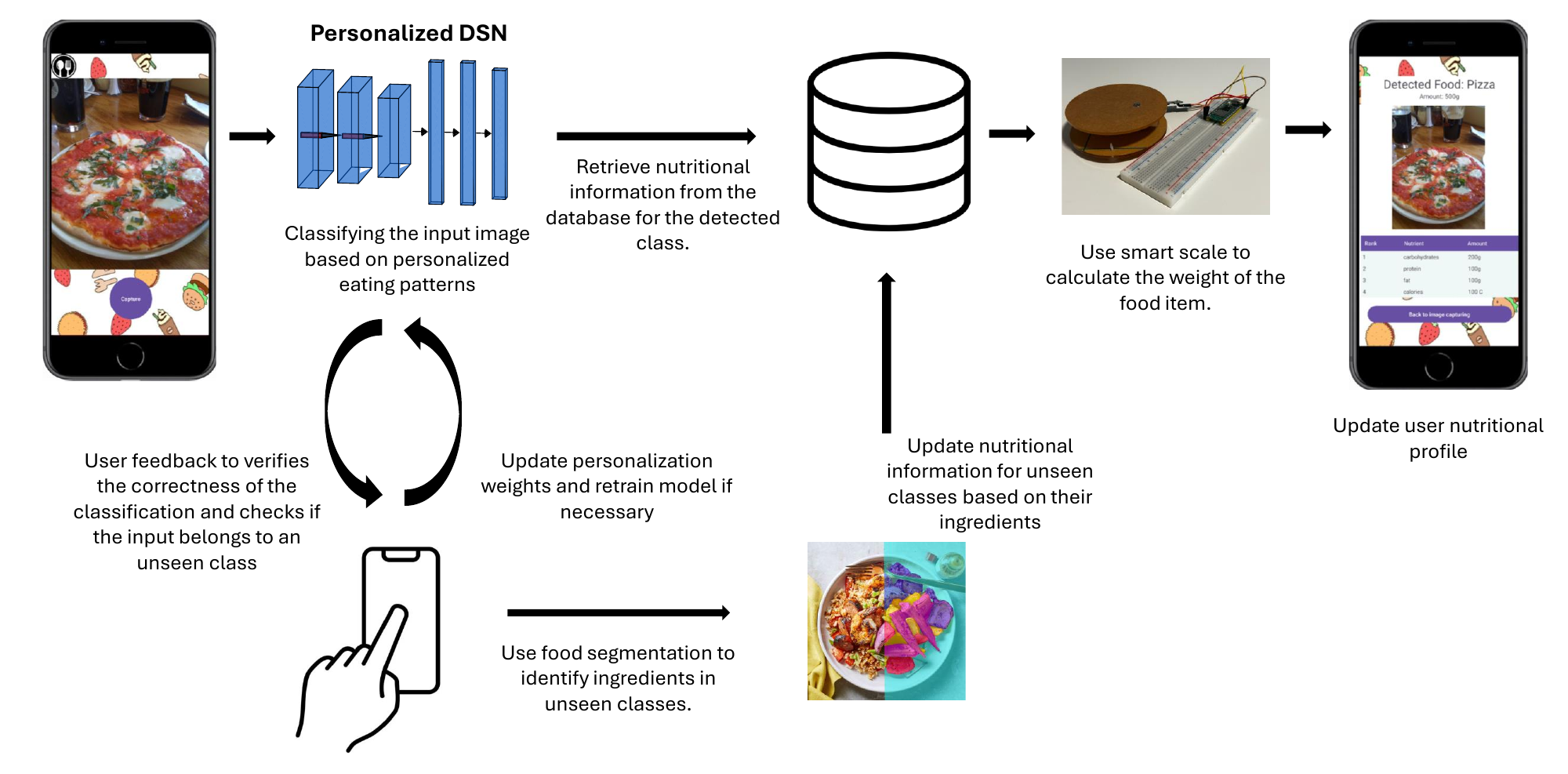}
\caption{Overview of our framework for food intake monitoring system leveraging PDSN.}
\label{Framework}
\vspace{-10pt}
\end{figure*}
Recently, with the outstanding achievements of transformers in the field of natural language processing (NLP) introduced by \cite{vaswani2017attention}, researchers were attracted to the use of these types of architecture in computer vision tasks. \cite{dosovitskiy2020image} proposed an architecture called Vision Transformer which treats image patches as tokens and processes them using a transformer model. Moreover, \cite{liu2021swin} introduced Swin Transformer, a hierarchical transformer utilizing shifted window mechanisms, designed to handle image inputs of various sizes and achieve strong performance across multiple vision tasks.
Ultimately, in the realm of food classification, various researchers have employed these types of architectures to effectively categorize different 
foods \cite{kim2024global}.

\subsection{Class incremental classification}
Continual learning, also known as incremental learning or lifelong learning is an area of machine learning that aims to learn continuously and adapt to new data and tasks over time without forgetting previous knowledge, which is a common issue known as catastrophic forgetting \cite{de2021continual}.

Continual learning can be categorized as instance-incremental learning (IIL), domain-incremental learning (DIL), task-incremental learning (TIL), class-incremental learning (CIL), task-free continual learning (TFCL), online continual learning (OCL), and continual pre-training (CPT) \cite{wang2024comprehensive}.

In class incremental learning, new classes appear over time and the model should detect new classes without losing the capability of classifying the ones. There are works focusing on using regularization-based approaches to achieve incremental 
learning \cite{lee2020continual}, while some other works have used
replay-based methods \cite{yang2022dynamic} and
parameter isolation methods \cite{xu2018reinforced}.

Moreover, in food classification, several works utilize continual learning to improve their food classification model further. \cite{he2023long} applied continual learning for a long-tailed food dataset and proposed a method to enhance the classification of class imbalance data. In another work, \cite{raghavan2024online} applied class incremental learning in an online learning setup, making their system more practical in real-world scenarios. 
\subsection{Personalized food classification}
The area of personalized food classification is relatively under-explored, presenting a compelling opportunity to advance food classification based on individual eating habits. In its early stages, \cite{wang2015use} proposed a personalized food image analysis model using recursive Bayesian estimation to learn individuals' eating habits. However, their model struggles with scalability and adaptability to new food classes due to the limitations of recursive Bayesian estimation in handling large and dynamic datasets. \cite{maruyama2010personalization} have also specifically explored the use of Bayesian networks as incremental learning networks. Furthermore, several research papers achieved personalized food classification by leveraging the nearest class mean classifier and the 1-nearest neighbor classifier 
\cite{yu2018food}. Finally, in more
recent approaches, \cite{pan2023personalized} utilized self-supervised learning to classify two new benchmark datasets, FOOD101-Personal and VFN-Personal, incorporating personalization. Despite their innovative approaches and promising results, all of these previous models focused solely on food images and sequences of food images, overlooking essential contextual information such as meal time, meal location, and meal frequency. Consequently, their accuracy was insufficient for practical use in real-world scenarios.

\section{Methodology and System Design}

\subsection{Personalized food intake monitoring framework}
Different from all existing work, this paper introduces a personalized food intake monitoring framework, as illustrated in Fig. \ref{Framework}. The core component of this framework is the personalized dynamic support network (PDSN), which classifies food images collected via our mobile application by integrating user profile history data, including meal frequency, meal time, and meal location.

Upon each food detection, user feedback regarding the accuracy of the detection is solicited, enabling continuous updates to our personalized module. This feedback mechanism allows the model to determine whether the detected food item is already included in the existing class set. If the food item is part of the existing classes, the system retrieves nutritional information from the nutrient database, including macronutrient content such as protein, carbohydrate, fat, and calorie values.

Subsequently, the mobile application interfaces with a smart scale via Bluetooth to ascertain the weight of the meal. By combining the meal weight with the nutritional data from the nutrient database, we accurately calculate the user's macronutrient intake for that particular meal.

In instances where the detected food item is not part of the existing class set, the continual learning capability of PDSN is employed to train the model to recognize new classes. Additionally, food image segmentation techniques are applied to identify the ingredients of the new food item, enabling the calculation of its macronutrient content. This new information is then incorporated into the nutrient database.

This personalized approach ensures that the monitoring framework adapts to individual eating habits and preferences, thereby providing accurate and tailored nutritional information to the user.

\subsection{Personalizer plug-in}

Existing personalized food classification models often utilize eating patterns as 
time series data to personalize the classifier \cite{pan2023personalized}. However, information such as 
meal frequency, meal time, and meal location are specific to the user and can be leveraged to achieve even greater accuracy and personalization of the model.
To utilize this information, we define the following:

Let $F$ be the set of existing food classes, $T$ be the discrete set of times for eating, and $L$ be the discrete set of locations where the user is eating.

For each user $u \in U$, we introduce a vector $\mathbf{MF}^{u}$ that holds the probability $P(F = f)$ of consuming a specific food, considering the frequency of appearance of that food in the user's profile history. This vector is denoted as $\mathbf{MF}^{u} \in \mathbb{R}^{|F| \times 1}$, where $\text{MF}^{u}_{f}$ represents the probability of consuming food type $f \in F$ for user $u$. The sum of $\text{MF}^{u}_{f}$ for $f$ in the range $0$ to $|F|$ is calculated as \begin{equation}
\sum_{f=0}^{|F|} \text{MF}^{u}_{f} = 1,
\label{eq:1}
\end{equation}
where $\text{MF}^{u}_{f}$ is uniformly distributed across all food classes at the initialization phase.

We define the matrix $\mathbf{MT}^{u} \in \mathbb{R}^{|F| \times |T|}$ that represents the conditional probability $P(T = t \mid F = f)$  of consuming a meal at a specific time $t \in T$ for a given food type, considering the user's profile history, where $\text{MT}^{u}_{f,t}$ represents the probability of consuming food type $f$ at time $t$ for user $u$. The sum of $\text{MT}^{u}_{f,t}$ for $t$ in the range $0$ to $|T|$ is calculated as
\begin{equation}
\sum_{t=0}^{|T|} \text{MT}^{u}_{f,t} = 1,
\label{eq:2}
\end{equation} where $\text{MT}^{u}_{f,t}$ is uniformly distributed across all times for each food class at the initialization phase.
\begin{figure}[t] 
\centering
\includegraphics[width=3in]{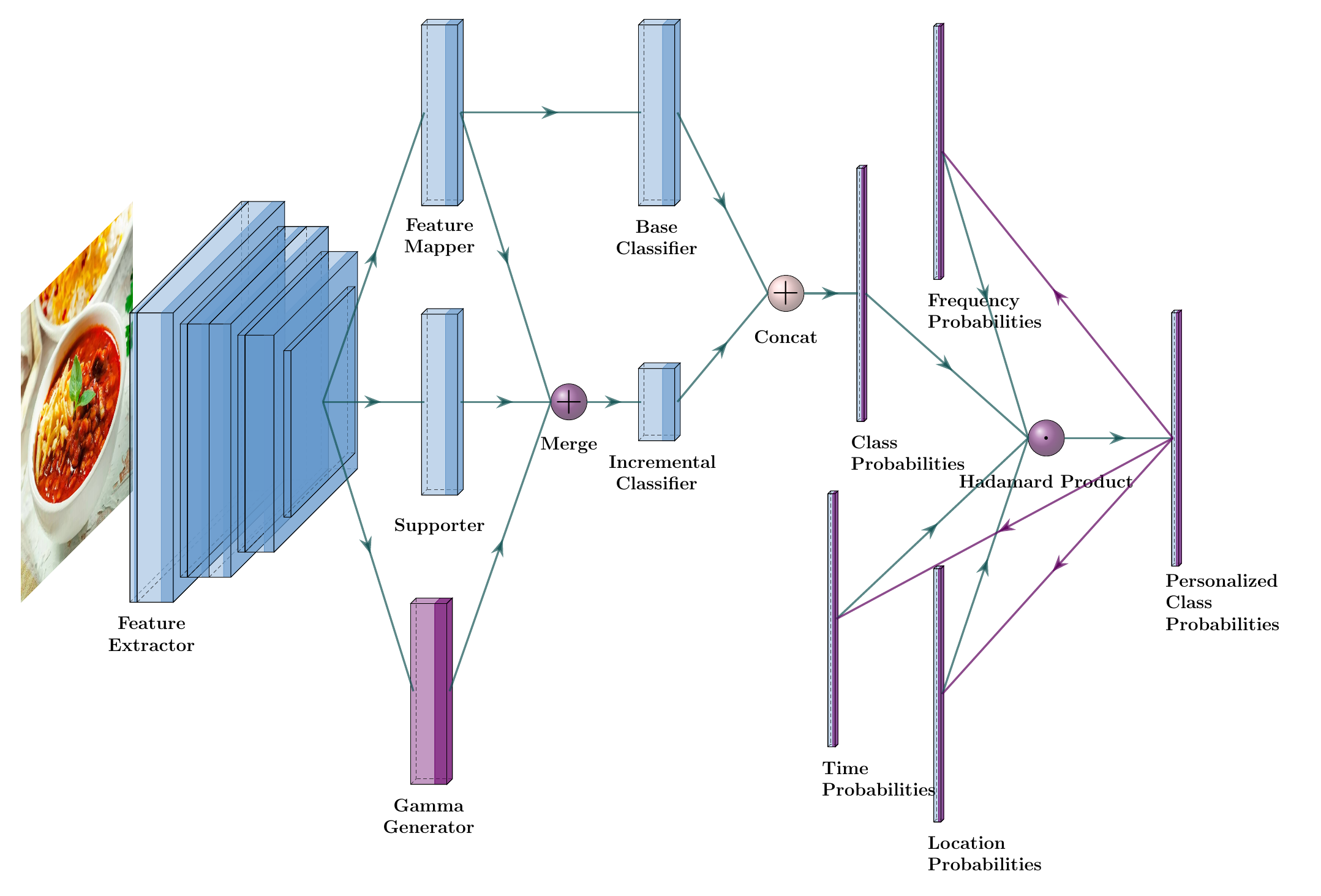}
\caption{Architecture of PDSN showcasing its innovative adaptive food image classification design.}
\label{DSN}
\end{figure}

We define the matrix $\mathbf{ML}^{u} \in \mathbb{R}^{|F| \times |L|}$ that represents the conditional probability $P(L = l \mid F = f)$ of consuming a meal at a specific location $l \in L$ for a given food type, considering the user's profile history, where $\text{ML}^{u}_{f,l}$ represents the probability of consuming food type $f$ at location $l$ for user $u$. The sum of $\text{ML}^{u}_{f,l}$ for $l$ in the range $0$ to $|L|$ is calculated as
\begin{equation}
\sum_{l=0}^{|L|} \text{ML}^{u}_{f,l} = 1,
\label{eq:3}
\end{equation}
where $\text{ML}^{u}_{f,l}$ is uniformly distributed across all locations for each food class at the initialization phase.

Upon receiving each input image $I \in \mathbb{R}^{n \times n}$, it undergoes classification processes using classifiers $C$ to generate a probability distribution $P$ across the classes. Specifically, $C(I) = P$ represents the transformation of input image $I$ to the probability distribution $P$. Let $P^{u}_{f}$ denote the probability of class $f$ for user $u$ based on the input image.

The personalized probability distribution $PP^{u}$ can be expressed as

\begin{equation}
PP^{u}_{f} = P^{u}_{f} \cdot \text{MF}^{u}_{f} \cdot \text{MT}^{u}_{f, t_i} \cdot \text{ML}^{u}_{f, l_i},
\end{equation}
where $u \in U$, $t_i \in T$, and $l_i \in L$ represent the user, time, and location associated with the input image, respectively.
The detected class is the one with the maximum value in the personalized probability distribution and expressed as
\begin{equation}
\text{Detected Class} = f_i = \arg\max_f (PP^{u}_{f}).
\end{equation}

After detecting the food type from the input image, if the model's prediction is incorrect, the correct class will be requested. Based on this correct class, the personalization vectors will then be updated as follows.

    Let $\alpha_f$, $\alpha_t$, and $\alpha_l$ be the forgetting factors for meal frequency, meal time, and meal location, respectively. These values will be used to update the personalization vectors in such a way that the probability of detecting foods more important and specific to the user will increase, while the probability of detecting less relevant foods will decrease by the forget factors.
   
    Update the food probability vector $\mathbf{MF}^{u}_{f}$, the time probability matrix $\mathbf{MT}^{u}_{f,t}$, and the location probability matrix $\mathbf{ML}^{u}_{f,l}$ as
     \begin{equation}
     \mathbf{MF}^{u}_{f} = \begin{cases}
       \mathbf{MF}^{u}_{f} + \alpha_f \cdot (1 - \mathbf{MF}^{u}_{f}), & \text{if } f = f_i, \\
       \mathbf{MF}^{u}_{f} \cdot (1 - \alpha_f), & \text{otherwise}.
     \end{cases}
     \end{equation}
     \begin{equation}
     \mathbf{MT}^{u}_{f,t} = \begin{cases}
       \mathbf{MT}^{u}_{f,t} + \alpha_t \cdot (1 - \mathbf{MT}^{u}_{f,t}), & f = f_i \text{ , } t = t_i, \\
      \mathbf{MT}^{u}_{f,t} \cdot (1 - \alpha_t), &  f = f_i \text{ , } t \neq t_i, \\
       \mathbf{MT}^{u}_{f,t}, & \text{otherwise}.
     \end{cases}
     \end{equation}
     \begin{equation}
     \mathbf{ML}^{u}_{f,l} = \begin{cases}
       \mathbf{ML}^{u}_{f,l} + \alpha_l \cdot (1 - \mathbf{ML}^{u}_{f,l}), & f = f_i \text{ , } l = l_i, \\
       \mathbf{ML}^{u}_{f,l} \cdot (1 - \alpha_l), & f = f_i \text{ , } l \neq l_i, \\
       \mathbf{ML}^{u}_{f,l}, & \text{otherwise}.
     \end{cases}
     \end{equation}
In the mentioned equations, the probability of detecting less relevant food types will be decreased by a forgetting factor, and this reduction will be reallocated to the probabilities of more relevant foods for the user.
Moreover, in this way, we ensure that equations \eqref{eq:1} to \eqref{eq:3} remain satisfied, and by incorporating these user-specific probabilities, our method aims to enhance the accuracy of personalized food classification models by leveraging detailed and individualized information on meal frequency, meal time, and meal location.

\subsection{Personalized dynamic support network}

In this section, we propose a modified version of DSN \cite{yang2022dynamic}. The overall architecture follows the DSN structure but introduces an improvement for personalized food classification through incremental learning, as illustrated in Fig. \ref{DSN}.

The proposed model consists of the following key components:

\subsubsection*{Backbone Feature Extraction}
The model utilizes a backbone feature extractor, such as ResNet \cite{he2016deep}, to extract features from the input images that can be used for classifying the input. The extracted features can be expressed as 
\begin{equation}
\mathbf{h} = \text{backbone}(x),
\end{equation}
where \(x\) is the input image, and \textit{backbone} represents the model used for feature extraction.

\subsubsection*{Feature Mapping}
The extracted features are mapped to a new feature space using a fully connected layer, converting the raw features into a form suitable for classification, as follows
\begin{equation}
\mathbf{z} = \mathbf{W}_{fm} \mathbf{h},
\end{equation}
where \(\mathbf{W}_{fm}\) is the weight matrix of the fully connected layer.
These features are then normalized to ensure consistent scale and improve the performance of the classification model, as follows
\begin{equation}
\mathbf{z} = \frac{\mathbf{z}}{\|\mathbf{z}\|_2}.
\end{equation}

\subsubsection*{Base Class Classification}
The normalized features are passed through another fully connected layer for the classification of base classes as follows
\begin{equation}
\text{output}_0 = \frac{\mathbf{W}_0 \mathbf{z}}{\|\mathbf{W}_0\|_2 \|\mathbf{z}\|_2},
\end{equation}
where \(\mathbf{W}_0\) is the weight matrix of the base classifier.

\subsubsection*{Gamma Generation for Incremental Sessions}
The output of the feature extracting layer is also passed to another fully connected neural network to generate gamma values for incremental sessions as follows
\begin{equation}
\gamma = \text{relu}(\mathbf{W}_{\gamma} \mathbf{h}),
\end{equation}
where \(\mathbf{W}_{\gamma}\) is the weight matrix of the gamma generator.

\subsubsection*{Session-specific Classifiers}
For each incremental session \(i \geq 1\), a support mechanism maps the features to the appropriate space for incremental classes. These mapped features are then merged with the features generated for the base classes to incorporate their information, regulated by the responsible \(\gamma\).  Finally, the output is calculated using the merged features as follows
\begin{equation}
\text{supporter}_i = \frac{\mathbf{W}_{s, i} \mathbf{h}}{\|\mathbf{W}_{s, i}\|_2 \|\mathbf{h}\|_2},
\end{equation}
\begin{equation}
\mathbf{z}_i = \gamma_{i} \mathbf{z} + \text{supporter}_i,
\end{equation}
\begin{equation}
\mathbf{z}_i = \frac{\mathbf{z}_i}{\|\mathbf{z}_i\|_2},
\end{equation}
\begin{equation}
\text{output}_i = \frac{\mathbf{W}_i \mathbf{z}_i}{\|\mathbf{W}_i\|_2 \|\mathbf{z}_i\|_2},
\end{equation}
where \(\mathbf{W}_{s, i}\) is the weight matrix of the supporter for session \(i\), and \(\mathbf{W}_i\) is the weight matrix of the classifier for session \(i\).

\subsubsection*{Concatenation of Outputs}
Probability distribution across the different classes is a concatenation of the outputs from all sessions obtained as
\begin{equation}
\text{output} = \left[ \text{output}_0 \|\ldots \ \| \, \text{output}_i \, \| \, \ldots \, \| \, \text{output}_{sess} \right]
\end{equation}
where $\text{output}_i$ represents the output at session $i$, $\|$ denotes the concatenation operation, and $sess$ denotes the total number of sessions.
\subsubsection*{Personalizer}
At the end, the personalizer plug-in will be used to incorporate contextual information such as meal frequency, meal time, and meal location further to enhance the model's performance for that specific user.

The major improvement of our approach over the original DSN is the personalized plug-in and network's dynamic generation of gamma. In the original DSN, gamma is treated as a hyperparameter requiring manual tuning. Instead, our approach enables the network to dynamically adjust the impact of the base class feature mapper and the incremental class feature mappers, thereby improving classification performance and adaptability to new data.

By learning gamma values, the network can better decide the relative importance of base and incremental class features for each specific data point, leading to more accurate and personalized food classification.

This methodology outlines the design and implementation of our personalized dynamic support network (PDSN) model highlighting the architectural innovations and the theoretical foundations supporting its improved performance in incremental learning scenarios.

\begin{table*}[!ht]
\centering
\caption{Average Top-1 Classification Accuracy $\pm$ std for FOOD101-Personal and VFN-Personal in different time steps}
\label{tab:accuracy}
\begin{tabular}{l|cccc||cccc}
\toprule
\textbf{Model} & \multicolumn{4}{c}{\textbf{FOOD101-Personal}} & \multicolumn{4}{c}{\textbf{VFN-Personal}} \\
\cmidrule(lr){2-5} \cmidrule(lr){6-9}
 & \textbf{t$_{75}$} & \textbf{t$_{150}$} & \textbf{t$_{225}$} & \textbf{t$_{300}$} & \textbf{t$_{75}$} & \textbf{t$_{150}$} & \textbf{t$_{225}$} & \textbf{t$_{300}$} \\
\midrule
AlexNet & 92.26 $\pm$ 0.03 & 92.16 $\pm$ 0.02 & 91.37 $\pm$ 0.02 & 91.35 $\pm$ 0.02 & 58.46 $\pm$ 0.08 & 58.74 $\pm$ 0.06 & 58.87 $\pm$ 0.07 & 58.51 $\pm$ 0.07 \\
VGGNet & 94.93 $\pm$ 0.02 & 94.80 $\pm$ 0.01 & 94.53 $\pm$ 0.01 & 94.53 $\pm$ 0.01 & 62.66 $\pm$ 0.07 & 62.71 $\pm$ 0.06 & 63.00 $\pm$ 0.06 & 62.91 $\pm$ 0.06 \\
GoogLeNet & 94.73 $\pm$ 0.02 & 94.33 $\pm$ 0.01 & 94.06 $\pm$ 0.01 & 93.75 $\pm$ 0.01 & 62.15 $\pm$ 0.07 & 62.58 $\pm$ 0.07 & 63.14 $\pm$ 0.08 & 63.03 $\pm$ 0.08 \\
ResNet & 94.60 $\pm$ 0.02 & 94.63 $\pm$ 0.01 & 94.37 $\pm$ 0.01 & 94.26 $\pm$ 0.01 & 62.46 $\pm$ 0.07 & 63.12 $\pm$ 0.07 & 62.90 $\pm$ 0.08 & 62.55 $\pm$ 0.07 \\
DenseNet & 95.53 $\pm$ 0.01 & 95.63 $\pm$ 0.01 & 95.35 $\pm$ 0.01 & 95.30 $\pm$ 0.01 & 67.94 $\pm$ 0.08 & 69.30 $\pm$ 0.07 & 69.55 $\pm$ 0.07 & 69.20 $\pm$ 0.07 \\
Vision Transformer & 97.46 $\pm$ 0.01 & 96.90 $\pm$ 0.01 & 96.88 $\pm$ 0.01 & 96.91 $\pm$ 0.01 & 71.94 $\pm$ 0.06 & 73.00 $\pm$ 0.07 & 73.43 $\pm$ 0.07 & 73.14 $\pm$ 0.07 \\
Swin Transformer & 97.46 $\pm$ 0.01 & 97.36 $\pm$ 0.01 & 96.95 $\pm$ 0.01 & 96.93 $\pm$ 0.01 & 70.97 $\pm$ 0.07 & 71.79 $\pm$ 0.07 & 71.91 $\pm$ 0.07 & 71.61 $\pm$ 0.06 \\
\midrule
 \textbf{(Our Work)} & & & & & & & & \\
\midrule
P-AlexNet & 92.86 $\pm$ 0.03 & 93.53 $\pm$ 0.02 & 93.55 $\pm$ 0.02 & 93.65 $\pm$ 0.02 & 62.71 $\pm$ 0.07 & 66.28 $\pm$ 0.08 & 68.88 $\pm$ 0.08 & 69.78 $\pm$ 0.08 \\
P-VGGNet & 95.46 $\pm$ 0.02 & 95.63 $\pm$ 0.01 & 95.71 $\pm$ 0.01 & 95.80 $\pm$ 0.01 & 68.87 $\pm$ 0.07 & 72.48 $\pm$ 0.07 & 74.58 $\pm$ 0.07 & 75.35 $\pm$ 0.07 \\
P-GoogLeNet & 95.66 $\pm$ 0.02 & 95.33 $\pm$ 0.01 & 95.51 $\pm$ 0.01 & 95.23 $\pm$ 0.00 & 67.69 $\pm$ 0.08 & 71.46 $\pm$ 0.08 & 73.77 $\pm$ 0.07 & 74.19 $\pm$ 0.07 \\
P-ResNet & 95.26 $\pm$ 0.02 & 95.63 $\pm$ 0.01 & 95.71 $\pm$ 0.01 & 95.46 $\pm$ 0.01 & 67.12 $\pm$ 0.07 & 71.69 $\pm$ 0.08 & 73.98 $\pm$ 0.07 & 74.60 $\pm$ 0.07 \\
P-DenseNet & 95.93 $\pm$ 0.01 & 96.46 $\pm$ 0.01 & 96.44 $\pm$ 0.01 & 96.38 $\pm$ 0.01 & 73.48 $\pm$ 0.08 & 76.58 $\pm$ 0.08 & 78.56 $\pm$ 0.07 & 78.69 $\pm$ 0.07 \\
P-Vision Transformer & 97.73 $\pm$ 0.01 & 97.53 $\pm$ 0.01 & 97.37 $\pm$ 0.01 & 97.35 $\pm$ 0.00 & 76.61 $\pm$ 0.06 & \textbf{79.76 $\pm$ 0.07} & \textbf{81.50 $\pm$ 0.06} & 81.66 $\pm$ 0.07 \\
P-Swin Transformer & \textbf{97.86 $\pm$ 0.01} & \textbf{98.06 $\pm$ 0.01} & \textbf{97.73 $\pm$ 0.01} & \textbf{97.45 $\pm$ 0.01} & \textbf{76.87 $\pm$ 0.07} & 79.64 $\pm$ 0.07 & 81.45 $\pm$ 0.06 & \textbf{81.74 $\pm$ 0.06} \\
\bottomrule
\end{tabular}
\vskip 1mm
\raggedright The "P-" prefix denotes the integration of our personalizer plug-in to the baseline models. \par
\vspace{-10pt}
\end{table*}

\begin{table}[!t]
\centering
\caption{Best Top-1 Accuracy of Classification Architectures on FOOD101 and VFN after 20 Epochs}
\label{tab:accuracy_20epochs}
\begin{tabular}{l||c|c}
\toprule
\textbf{Model} & \textbf{FOOD101} & \textbf{VFN} \\

\midrule
AlexNet & 64.49 & 86.07 \\
VGGNet & 76.65 & \textbf{86.46} \\
GoogLeNet & 75.74 & 86.34 \\
ResNet& 75.08 & 86.04 \\
DenseNet & 81.80 & 86.07 \\
Vision Transformer & 84.37 & 86.34 \\
Swin Transformer& \textbf{87.86} & 85.75 \\
\bottomrule
\end{tabular}
\vspace{-10pt}
\end{table}

\begin{table*}[!t]
\centering
\caption{Accuracy Breakdown of DSN and PDSN Models on Different Datasets}
\label{tab:accuracy-breakdown}
\begin{tabular}{l|cccc||cccc||cccc||cccc}
\toprule
\textbf{Model} & \multicolumn{4}{c||}{\textbf{FOOD101}} & \multicolumn{4}{c||}{\textbf{FOOD101-Personal}} & \multicolumn{4}{c||}{\textbf{VFN}} & \multicolumn{4}{c}{\textbf{VFN-Personal}} \\
\cmidrule(lr){2-5} \cmidrule(lr){6-9} \cmidrule(lr){10-13} \cmidrule(lr){14-17}
 & \textbf{Base} & \textbf{New} & \textbf{Total} &  & \textbf{Base} & \textbf{New} & \textbf{Total} &  & \textbf{Base} & \textbf{New} & \textbf{Total} &  & \textbf{Base} & \textbf{New} & \textbf{Total} &  \\
\midrule
DSN & \textbf{74.85} & 73.33 & \textbf{74.82} &  & 94.26 & 72.42 & 92.28 &  & 85.95 & 70.00 & 85.57 &  & 62.54 & 69.74 & 63.20 &  \\
PDSN (Our Work) & 74.73 & \textbf{76.66} & 74.77 &  & \textbf{95.26} & \textbf{77.54} & \textbf{93.65} &  & \textbf{85.98} & \textbf{70.00} & \textbf{85.59} &  & \textbf{74.05} & \textbf{70.40} & \textbf{73.71} &  \\
\bottomrule
\end{tabular}
\vspace{-10pt}
\end{table*}

\section{Experiments and Results}
In this section, we evaluate the effectiveness of our personalized food intake monitoring model. The objective is to demonstrate the enhanced accuracy of food classification in a practical setting, achieved by incorporating user-specific information such as meal frequency, meal time, and meal location. We compare the performance of different classification models, as discussed in Section \ref{sec:related}, both with and without our personalization method. Additionally, we demonstrate the superior effectiveness of our approach compared to the original DSN \cite{yang2022dynamic} in incrementally classifying user-specific food categories. Finally, we conduct an ablation study to examine the impact of various user-specific information (meal frequency, meal time, and meal location) on classification accuracy.
\subsection{Experimental setup}

\subsubsection{Dataset}
Our experiment utilized two primary datasets: Food-101 and VFN. Additionally, we integrated personalized versions of these datasets, referred to as Food-101-Personal and VFN-Personal, for evaluating personalized food image classification tasks.

The Food-101 dataset \cite{bossard2014food} is a widely used benchmark for food image classification, comprising 101,000 food images, with each class having 750 training images. It covers a wide spectrum of food categories essential for training machine learning models in food image classification tasks.

The Food-101-Personal dataset was derived through an online survey conducted by \cite{pan2023personalized} using the Food-101 dataset. Participants simulated one week of food consumption patterns by selecting foods from the FOOD101 categories. This personalized dataset includes 20 patterns with an average of 44 food classes and 6000 images in total. It comprises 300 images per individual, representing long-term food consumption patterns for enhanced evaluation in personalized food image classification tasks.

The VFN dataset \cite{mao2021visual} comprises 14,991 online food images sourced from the What We Eat In America (WWEIA) database \cite{usda_wweia}. The dataset is designed specifically for food recognition and includes images from various food categories. The characteristics of VFN make it highly relevant for evaluating the generalizability of our models. This dataset covers 82 food categories selected based on high intake frequency from the WWEIA database, reflecting the most commonly consumed foods in the United States.

The VFN-Personal dataset originated from a dietary examination conducted by \cite{pan2023personalized} involving healthy individuals aged 18 to 65, employing an image-based dietary evaluation approach. Participants documented their food intake over a three-day period. Much like the Food-101-Personal dataset, the VFN-Personal dataset was crafted utilizing techniques to mimic prolonged food consumption behaviors. The dataset comprises 26 patterns, each encompassing an average of 29 distinct classes and similar to the Food-101-Personal dataset, it includes 300 images per individual.

\subsubsection{Baselines}
Our study centers on assessing the performance of seven widely recognized architectures known for their image classification capabilities: AlexNet \cite{krizhevsky2012imagenet}, VGGNet \cite{simonyan2014very}, GoogLeNet \cite{szegedy2015going}, ResNet \cite{he2016deep}, DenseNet \cite{huang2017densely}, Vision Transformer \cite{dosovitskiy2020image}, and Swin Transformer \cite{liu2021swin}.

We selected these architectures based on their track record of high performance in image classification tasks. To ensure a fair comparison, we utilized pretrained versions of these models trained on the ImageNet \cite{deng2009imagenet} dataset. Fine-tuning was performed specifically for food image classification, involving 20 epochs of training to adapt the models to our dataset.

We conducted a comprehensive comparison between the baseline models and their enhanced versions. The enhancements include the incorporation of our proposed personalized plugin, designed to improve classification accuracy by tailoring models to individual eating habits.

To assess the effectiveness of our model in learning new classes incrementally, we compared DSN \cite{yang2022dynamic} and PDSN architectures based on the accuracy of different datasets using the same backbone to ensure a fair comparison. This setup allowed us to study how our approach facilitates better learning of new classes over time.

\subsubsection{Implementation Details}

We utilized various pretrained architectures implemented in Python with the PyTorch \cite{paszke2019pytorch} library as the backbones of DSN and PDSN to extract features. Each architecture was fine-tuned for food image classification tasks with a batch size of 32 and trained for 20 epochs.

Input images were resized to \(224 \times 224\) pixels to accommodate the requirements of different architectures. For optimization, we employed the Stochastic Gradient Descent (SGD) optimizer with a learning rate (\(lr\)) of 0.001, momentum of 0.9, weight decay of 0.0005, and nesterov momentum 
 enabled.

To incorporate meal time and location information, each input image had a fixed probability of being associated with different meal times and locations based on the type of food. This probabilistic approach introduced variability in meal times and locations, mitigating potential overfitting. Also forget factors, \(\alpha_f\), \(\alpha_t\), and \(\alpha_l\) were set to 0.003, 0.04, and 0.04, respectively.

We trained the models on the FOOD101 and VFN datasets and subsequently evaluated their performance using the FOOD101-Personal and VFN-Personal datasets, which contain personalized eating patterns across 300 meal sessions.

This approach allowed us to assess the effectiveness of our personalized food classification system under realistic consumption scenarios, considering both temporal and spatial meal variations.

\subsection{Experimental Results}
Table \ref{tab:accuracy} presents the comparative performance metrics of our PDSN model alongside seven baseline architectures for food classification using the FOOD101-Personal and VFN-Personal datasets, both with and without our personalizer plug-in.

Initially, we trained all baseline models—AlexNet, VGGNet, GoogLeNet, ResNet, DenseNet, Vision Transformer, and Swin Transformer—on the standard FOOD101 and VFN datasets. Table \ref{tab:accuracy_20epochs} summarizes the accuracy results of these baseline models after training them for 20 epochs on FOOD101 and VFN. After evaluating the baseline models, we integrated our personalizer plug-in into each architecture and re-evaluated their performance on the personalized versions of these datasets, FOOD101-Personal and VFN-Personal, with and without enabling personalization.

FOOD101, being a well-populated dataset, demonstrated moderate improvements with the personalizer plug-in due to its ample data availability. Conversely, the VFN dataset, which lacks extensive data and initially had lower baseline model accuracy, benefited substantially from the personalizer plug-in. In some instances, we observed performance boosts of up to 10\% in classification accuracy.

Our personalizer plug-in significantly improved the models' performance by customizing the classification process according to individual eating patterns. This personalization leverages unique patterns in a user's food consumption, allowing the model to focus on the most relevant foods for each individual. By integrating personalization data into the classification process, our approach effectively reduces the negative effects of intra-class diversity and inter-class similarity by considering additional factors specific to each user's eating habits. This tailored learning process helps the models differentiate between similar food images in different classes and distinguish diverse food images within a same class more effectively.

To demonstrate the efficacy of our PDSN model in incremental learning scenarios, we conducted experiments, comparing to the original DSN architecture. Both models were initially trained on the complete set of classes from FOOD101 and VFN datasets separately, serving as the base session. Subsequently, we introduced two new classes to each dataset and evaluated their performance on four datasets: FOOD101, VFN, FOOD101-Personal, and VFN-Personal.

Table \ref{tab:accuracy-breakdown} summarizes the results of our incremental learning experiment. In standard datasets (FOOD101 and VFN), the improvement with PDSN was marginal, indicating that it learned new classes slightly better than DSN. This improvement can be attributed to the dynamic gamma generation in PDSN, which allows the model to effectively balance and integrate information from the base classifier when learning new classes.

In personalized versions (FOOD101-Personal and VFN-Personal), where our personalizer plug-in enhances adaptability, the performance improvement was notable. Specifically, PDSN achieved approximately a 5\% improvement in detecting new classes in FOOD101-Personal and an overall accuracy improvement of about 10\% in VFN-Personal compared to DSN.

This experiment underscores the effectiveness of our model's personalized approach, particularly in scenarios with limited data (VFN dataset) and for tasks requiring robust adaptation to new information.

\subsection{Ablation Study}

\begin{figure}[h]
\vspace{-15pt}
    \centering
    \includegraphics[width=\linewidth]{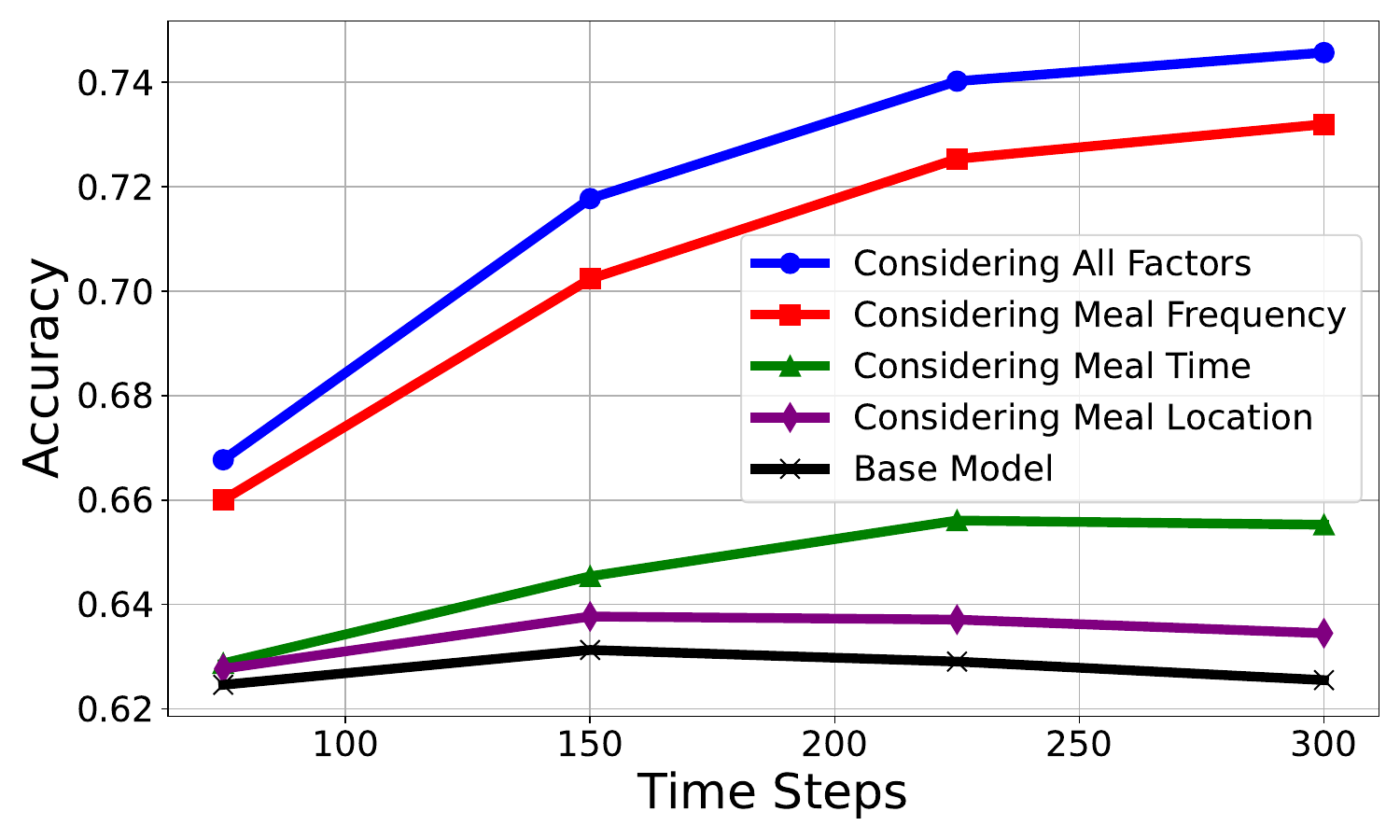}
    \caption{Accuracy over Time Steps Considering Different Factors}
    \label{fig:accuracy_plot}
    \vspace{-10pt}
\end{figure}

To investigate the impact of different meal-related factors (meal frequency, meal time, and meal location) on personalization and performance improvement, we conducted an ablation study using our PDSN model. Fig. \ref{fig:accuracy_plot} illustrates the performance of five scenarios over time, each evaluating different configurations of these factors.

The experiments were structured as follows. We first evaluated the base model without considering any meal factors. Subsequently, we separately assessed each factor's impact (meal frequency, meal time, meal location). Finally, we evaluated the model's performance when considering all factors simultaneously.

As depicted in Fig. \ref{fig:accuracy_plot}, integrating all meal factors into the personalization process yielded the best overall performance improvement. Specifically:
\begin{itemize}
    \item considering all meal factors collectively resulted in the highest performance gains, indicating the synergistic effect of comprehensive personalization.
Meal frequency had the most significant individual impact on performance improvement, demonstrating its critical role in enhancing the model's adaptability.
    \item meal time also contributed significantly to performance enhancement, although to a lesser extent compared to meal frequency.
    \item meal location, while contributing to improvement, had the least pronounced effect among the factors studied.
    \item the base model, which did not consider any meal factors, exhibited the lowest performance improvement, highlighting the necessity of personalized adaptation in food intake monitoring scenarios.
\end{itemize}

This ablation study underscores the importance of integrating contextual meal information into the model's learning process, emphasizing the role of meal frequency as a key determinant in enhancing classification accuracy and adaptability.
\section{Conclusion}

In this study, we introduced PDSN, a novel method tailored for personalized food intake monitoring systems. Our method mitigated the adverse impacts of both intra-class variation and inter-class similarities by incorporating user-specific factors related to individual eating behaviors. Moreover, our approach leverages dynamic gamma generation through the network, allowing for adaptive feature weighting between base and incremental classes. These innovations address the limitations of traditional DSN by enhancing classification accuracy and adaptability in the context of evolving dietary habits.

Through comprehensive experimentation, we evaluated PDSN with different baseline architectures on the FOOD101 and VFN datasets, demonstrating consistent performance improvements across personalized datasets (FOOD101-Personal and VFN-Personal). Our results underscored the effectiveness of the personalizer plug-in in enhancing classification accuracy, particularly in scenarios with limited data (VFN), where we observed up to a 10\% improvement over baseline models.

Furthermore, our study showcased PDSN's capability in incremental learning, where it outperformed traditional DSN by effectively classifying new food classes introduced after initial training sessions. This capability was most pronounced in personalized datasets, highlighting the model's robustness in adapting to individual dietary preferences over time.

Additionally, our ablation study highlighted the pivotal role of meal-related factors (meal frequency, meal time, and meal location) in further enhancing PDSN's performance. Integrating these factors significantly improved classification accuracy, with meal frequency proving to be the most influential factor in the personalization process.

In conclusion, PDSN represents a significant advancement in personalized food intake monitoring systems, offering robust performance improvements through dynamic feature adaptation and contextual meal information integration. Our findings not only contribute to the field of machine learning-driven dietary assessment but also pave the way for future research in personalized AI-driven healthcare applications.

\ifCLASSOPTIONcaptionsoff
  \newpage
\fi

\bibliographystyle{IEEEtran}
\bibliography{bibtex/bib/IEEEexample}
\end{document}